%% file: main.tex
\documentclass[10pt,twocolumn,letterpaper]{article}

\usepackage{cvpr}              %

\input{preamble}
\definecolor{cvprblue}{rgb}{0.21,0.49,0.74}
\usepackage[pagebackref,breaklinks,colorlinks,citecolor=cvprblue]{hyperref}
\newcommand{\bm}[1]{$#1$}

\def\paperID{31} %
\def\confName{CVPRW}
\def\confYear{2025}

\definecolor{darkgreen}{rgb}{0.0, 0.5, 0.0}

\newcommand{\sensitive}[1]{#1}

\newcommand{\drivesim}[0]{NVIDIA DRIVE Sim™}

\title{Data Scaling Laws for End-to-End Autonomous Driving}

\author{
Alexander Naumann\thanks{These authors contributed equally to this work conducted during internships at NVIDIA.}\textsuperscript{~~1} \and
Xunjiang Gu\textsuperscript{*2} \and
Tolga Dimlioglu\textsuperscript{*3} \and
Mariusz Bojarski\textsuperscript{1} \and
Alperen Degirmenci\textsuperscript{1} \and
Alexander Popov\textsuperscript{1} \and
Devansh Bisla\textsuperscript{1} \and
Marco Pavone\textsuperscript{1,4} \and
~~~~Urs Müller\thanks{These authors share senior authorship.}\textsuperscript{†1} \hspace{1.4cm}
Boris Ivanovic\textsuperscript{†1} \\
\\
\textsuperscript{1}NVIDIA \hspace{1cm}
\textsuperscript{2}University of Toronto \hspace{1cm}
\textsuperscript{3}New York University \hspace{1cm}
\textsuperscript{4}Stanford University
}

\begin{document}
\maketitle

\input{sec/0_abstract}

\input{sec/1_intro}

\input{sec/2_related_work}

\input{sec/3_dataset}

\input{sec/4_methods}

\input{sec/5_experiments}

\input{sec/6_conclusion}

{
    \small\bibliographystyle{ieeenat_fullname}
    \bibliography{main.bbl}
}

\input{sec/X_suppl}

\end{document}

%% file: sec/0_abstract.tex
\begin{abstract}

Autonomous vehicle (AV) stacks have traditionally relied on decomposed approaches, with separate modules handling perception, prediction, and planning. However, this design introduces information loss during inter-module communication, increases computational overhead, and can lead to compounding errors. To address these challenges, recent works have proposed architectures that integrate all components into an end-to-end differentiable model, enabling holistic system optimization. This shift emphasizes data engineering over software integration, offering the potential to enhance system performance by simply scaling up training resources. In this work, we evaluate the performance of a simple end-to-end driving architecture on internal driving datasets ranging in size from \sensitive{16 to 8192 hours} with both open-loop metrics and closed-loop simulations. Specifically, we investigate how much additional training data is needed to achieve a target performance gain, e.g., a 5\% improvement in motion prediction accuracy. By understanding the relationship between model performance and training dataset size, we aim to provide insights for data-driven decision-making in autonomous driving development.

\end{abstract}

%% file: sec/1_intro.tex
\vspace{-0.2cm}
\section{Introduction}
\label{sec:intro}

\begin{figure}[t]
    \centering
    \includegraphics[width=0.85\linewidth]{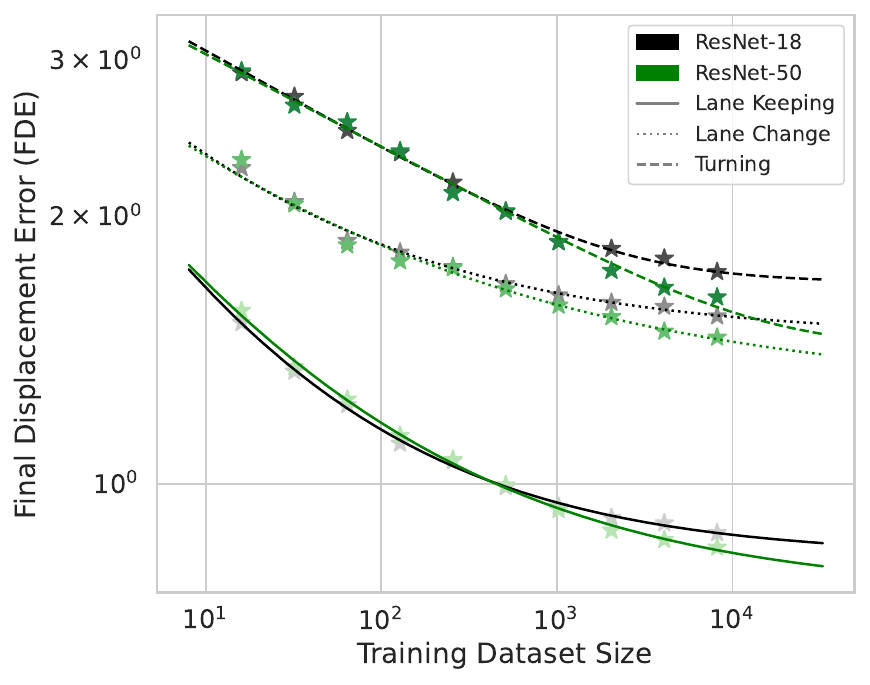}
    \vspace{-0.1cm}
    \caption{
        Scaling laws across different action types (lane keeping, lane change and turning) and model architectures (ResNet-18~\cite{HeZhangEtAl2016} and ResNet-50 backbone). 
        Note that larger models achieve faster performance gains across all scenarios in the large-data regime.
    }
    \vspace{-0.5cm}
    \label{fig:intro}
\end{figure}

Traditionally, autonomous driving systems have followed a modular approach, separating perception, prediction, and planning into distinct components, each optimized independently with their own objectives~\cite{MapTR, maptrv2, SalzmannIvanovicEtAl2020,Phan-MinhGrigoreEtAl2020, gao2020vectornet,liang2020lanegcn}. 
While this modular design distributes the effort of AV development across specialized teams, it introduces substantial challenges during integration, such as information loss between modules, compounding errors, and inefficient resource use. To address these limitations, end-to-end architectures have recently been proposed. These models take in sensor inputs, such as LiDAR and camera images, and directly output the planned driving path, integrating multiple modules into a single framework
where all components are jointly optimized toward a unified objective: motion planning. This unified optimization approach offers the potential to improve system performance by scaling up training resources. However, a critical question remains: how much training data is required to achieve meaningful performance gains in end-to-end autonomous driving systems?

In Natural Language Processing (NLP), numerous studies have examined scaling laws in terms of training data size and model performance~\cite{kaplan2020scaling, alabdulmohsin2022revisiting, zhai2022scaling}. In autonomous driving, however, the relatively limited size of public datasets has prevented similar large-scale analyses. This gap leaves uncertainty about how scaling impacts both the open-loop and, more critically, closed-loop performance of learning-based AV systems. Furthermore, it is unclear whether scaling specific components (e.g. perception or prediction) of end-to-end driving architectures yields significant performance improvements. This lack of clarity poses challenges as collecting and annotating autonomous driving data is costly. Establishing a reliable scaling law could allow AV developers to save substantial resources by better aligning data and training investments with performance gains.

Towards this end, we examine data scaling laws for AV models using a representative end-to-end driving stack, similar to those in earlier works such as~\cite{BojarskiDelTestaEtAl2016} and~\cite{codevilla2018end}, that employs imitation learning to directly map visual inputs (i.e. RGB camera images) to trajectories.
Concretely, we train and evaluate the model on internal datasets ranging from \sensitive{16 to 8192} hours of driving data, using standard open-loop metrics. We then assess its performance in a closed-loop simulator, where a rule-based controller converts the predicted trajectories into control commands.

Our contributions are fourfold: \textbf{First}, we present the first comprehensive and systematic analysis of scaling laws in end-to-end autonomous driving, as shown in \cref{fig:intro}. \textbf{Second}, we examine the challenges of training on large-scale datasets and propose a custom training scheme. \textbf{Third}, we investigate the amount of additional training data needed to achieve target performance gains, using various scaling law estimators. \textbf{Finally}, we integrate these trained models into the \drivesim{} closed-loop simulator\footnote{See \href{https://developer.nvidia.com/drive/simulation}{https://developer.nvidia.com/drive/simulation}.} to evaluate performance on key metrics, such as mean distance between failures (MDBF), that much more closely reflect real-world driving capability.

%% file: sec/2_related_work.tex
\section{Related Work}
\label{sec:related_work}

\subsection{Modular Driving Architectures}
Conventional autonomous driving systems employ a modular, cascaded architecture that separates perception, prediction, and planning. Accurate perception enables reliable environment understanding for autonomous driving, either through LiDAR~\cite{chen2023voxelnext, graham20183d, mao2021voxel, zhou2018voxelnet} or lower-cost vision-based alternatives~\cite{huang2023tri, li2023bevdepth, li2022bevformer, philion2020lss}. Recent advancements in RGB camera-based approaches have achieved competitive results across 3D object detection~\cite{reading2021categorical, zhang2022beverse, wang2022detr3d, huang2022bevdet}, HD map construction~\cite{MapTR, maptrv2, yuan2024streammapnet, liu2022vectormapnet, liu2024mgmap}, and 3D semantic occupancy prediction~\cite{tian2024occ3d, tong2023scene, wei2023surroundocc, huang2023tri}.

Building on this, the outputs of perception tasks serve as critical inputs for downstream modules, such as trajectory prediction and planning. Traditional trajectory prediction models use rasterized maps combined with agents' historical trajectories to forecast motion~\cite{SalzmannIvanovicEtAl2020,Phan-MinhGrigoreEtAl2020,YuanWengEtAl2021,GillesSabatiniEtAl2021,IvanovicHarrisonEtAl2023}. More recently, vectorized representations have improved deployment efficiency, with GNNs and Transformers encoding polyline maps and agent states~\cite{gao2020vectornet,ZhaoGaoEtAl2020,GillesSabatiniEtAl2022a,GillesSabatiniEtAl2022b,GuSunEtAl2021,zhou2022hivt,zhou2023query}. Although integration efforts between perception and prediction are advancing~\cite{GuSongEtAl2024, GuSongEtAl2024ECCV}, modular systems remain prone to compounding errors, information bottlenecks, and inefficiencies due to handcrafted filters and misaligned task objectives across modules.

\subsection{End-to-End Driving Architectures}

To address the challenges of modular systems, many end-to-end architectures have been proposed. Early approaches, such as~\cite{BojarskiDelTestaEtAl2016},~\cite{codevilla2018end} and~\cite{ALVINN}, relied solely on RGB camera inputs to directly learn vehicle control commands, bypassing intermediate stages such as perception and motion prediction. These methods are appealing due to their efficient runtime and elimination of information bottlenecks. More recent works, such as Transfuser~\cite{chitta2022transfuser}, incorporate additional sensor data like LiDAR, while TCP~\cite{wu2022trajectory} utilizes navigational commands to guide the generation of trajectories and controls. 
Despite their simplicity, such fully end-to-end methods have faced concerns for their lack of interpretability and safety, leading to the emergence of hybrid end-to-end approaches that retain certain modular elements~\cite{hu2023uniad, jiang2023vad, tong2023occnet, zheng2024genad, weng2024drive}. UniAD~\cite{hu2023uniad} integrates information from various preceding tasks, such as tracking, mapping, and prediction, achieving strong performance by jointly optimizing all tasks. VAD~\cite{jiang2023vad} improves the pipeline further by employing a vectorized scene representation,  reducing the computational overhead. PARA-Drive~\cite{weng2024drive} introduces a parallel architecture instead of a traditional serial pipeline. These approaches aim to optimize the entire system around a single, unified objective. By enabling the joint training of sub-components, this data-driven optimization has the potential to enhance system performance by simply scaling training resources—a key motivation for our work.

Although recent literature favors hybrid designs, we adopt a fully end-to-end approach in this paper. Specifically, we directly input camera images along with other relevant information, bypassing intermediate modules such as tracking and mapping. Rather than aiming for state-of-the-art performance, our work seeks to establish a systematic framework for analyzing the scaling laws of end-to-end driving architectures.%

\subsection{Scaling Laws and Data Estimation}

Scaling laws in deep learning have demonstrated predictable improvements in model performance as dataset sizes increase, following a power-law relationship~\cite{hestness2017deep, kaplan2020scaling, zhai2022scaling, rosenfeld2019constructive, henighan2020scaling, sun2017revisiting}: \( L_{\text{val}} \propto \beta x^c \), where \( x \) is the training data size, \( \beta \) and \( c \) are problem-specific constants, and \( L_{\text{val}} \) represents the validation loss. Studies such as~\cite{kaplan2020scaling} established that auto-regressive models like GPT exhibit consistent scaling with respect to dataset size, model capacity, and training iterations, insights later applied to optimize GPT-3's training~\cite{mann2020language}. Scaling laws have been observed across diverse modalities, from language and vision to multi-modal generative models~\cite{henighan2020scaling, zhai2022scaling}. Recent theoretical work further formalizes these observations by linking scaling behaviors to intrinsic data properties~\cite{sharma2022scaling, bisla2021theoretical, bahri2024explaining}.

A critical aspect of scaling laws is estimating the amount of additional data needed to achieve target performance, especially for costly data domains such as AVs~\cite{alabdulmohsin2022revisiting, mahmood2022much, mahmood2022optimizing, harvey2023probabilistic, jain2023meta, sasaki2024key, clark2024training}. \cite{mahmood2022much} introduced an active learning framework where estimators iteratively refine dataset size predictions, compensating for overly optimistic projections on smaller datasets. More recent approaches, including probabilistic models~\cite{harvey2023probabilistic} and meta-learning~\cite{jain2023meta}, aim to reduce extrapolation errors by distinguishing between data regimes. Notably, \cite{alabdulmohsin2022revisiting} proposes a generalized estimator to improve data requirement predictions across diverse tasks. Although some prior works in autonomous driving, such as EMMA~\cite{hwang2024emma}, STR~\cite{sun2023large}, and GUMP~\cite{hu2025solving}, briefly mention scaling laws as part of their analyses, they lack a systematic approach and do not derive a formal scaling relationship. 

A recent study~\cite{zheng2024preliminarydatascaling} explores data scaling in end-to-end AV systems but lacks a structured framework, omitting key analyses such as different scaling trends, performance variations across action types, and model capacity effects. In contrast, our work establishes a systematic approach for measuring scaling behavior and data requirement estimates. %

%% file: sec/3_dataset.tex
\section{Dataset Curation}
\label{sec:dataset}
\begin{figure}[htbp]
    \centering
    \includegraphics[width=0.8\linewidth]{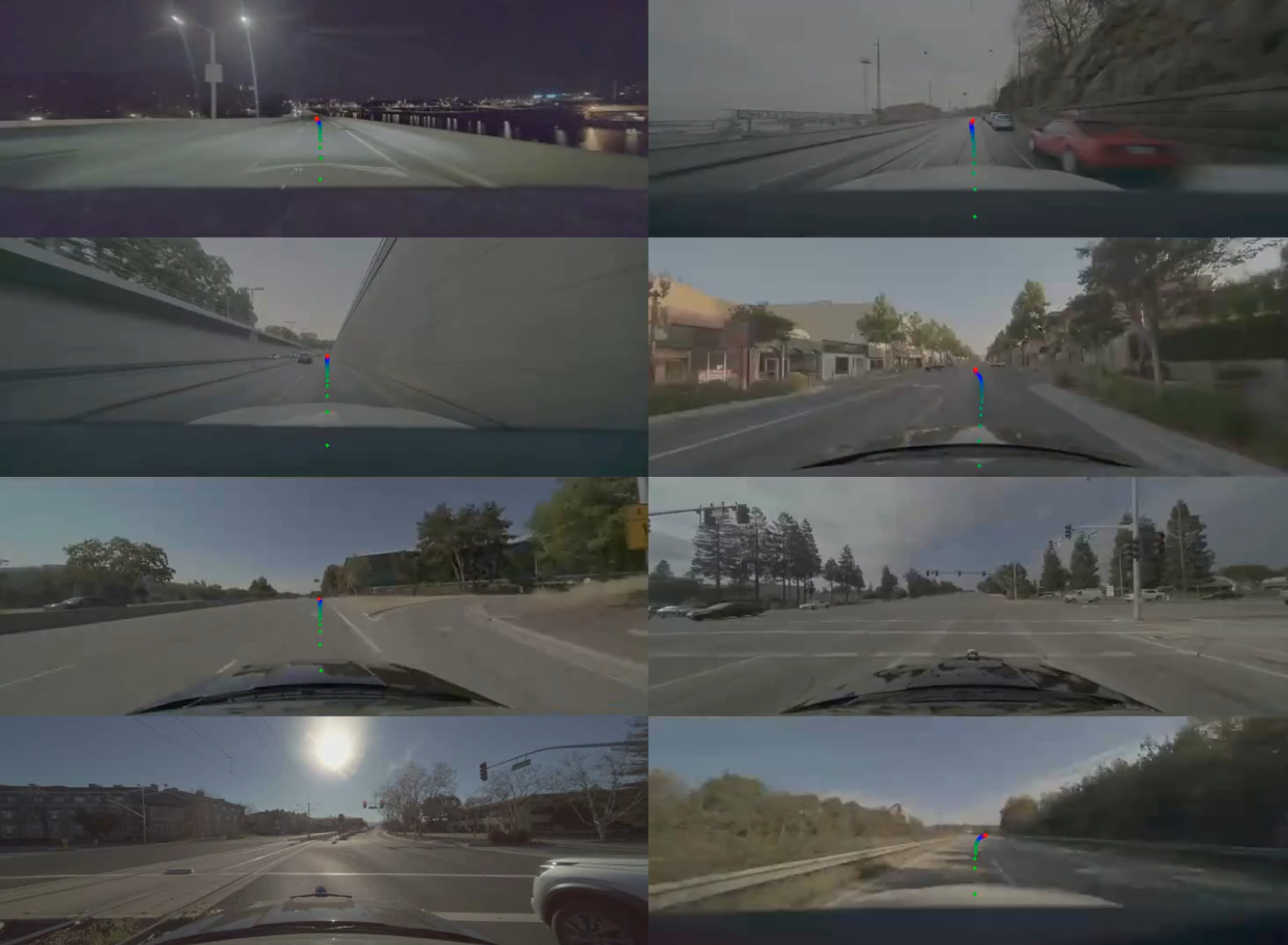}
    \caption{
        \sensitive{Randomly selected example images from the front camera with ground truth ego trajectory overlaid.}
    }
    \vspace{-5mm}
    \label{fig:dataset:ex}
\end{figure}

\vspace{-0.02cm}
To analyze scaling laws for end-to-end autonomous driving, we curate an industry-scale dataset \sensitive{of over 8,000 hours in duration and over 400,000~km in driven distance} from internal driving data across \sensitive{more than 10 countries}. 
The dataset provides 3 rectified images from \sensitive{3 wide-angle} cameras facing forward, to the left, and to the right of the AV, \sensitive{with 
a downscaled resolution of $734 \times 270$} collected at \sensitive{10 Hz}, as seen in \cref{fig:dataset:ex}.
The ego trajectory is uniformly resampled at \sensitive{5 Hz for 3 seconds} into the future.
To evaluate scaling laws, we prepare training datasets of varying sizes, ranging in powers of 2 from 16 hours to 8192 hours. 
This approach aligns our dataset scales with those of popular public benchmarks such as nuScenes \cite{CaesarBankitiEtAl2019} (15h), WOMD \cite{waymo_open_dataset} (574h), Lyft \cite{HoustonZuidhofEtAl2020} (1001h), and nuPlan \cite{nuplan} (1282h), enabling relevant comparisons in autonomous driving research.

\subsection{Geofencing}
\label{sec:dataset:geo}
To prevent information leakage across data partitions, we establish mutually-exclusive geographical regions, primarily across Europe and North America, for our training, validation, and test datasets, ensuring complete separation of spatial data.
We achieve this by constructing a global undirected graph where driving \textit{sessions} and visited H3 cells are represented as nodes, with edges connecting sessions to their respective H3 nodes. 
By identifying connected components within this graph, we determine \textit{session clusters}, i.e. geographically disjoint groups of sessions suitable for training, validation, or testing without risks of overlap.
Note that these groups of driving sessions can be of significant size, which makes proper allocation to our desired distribution of 96\% for training, 2\% for validation, and 2\% for testing challenging.
Further, while there is only one validation and test split, we analyze scaling laws by training on iteratively larger training datasets.
To prevent bias in the data selection, training split versions are \textit{cumulative}, meaning that the 16h training dataset is fully contained within the 32h training dataset.

\subsection{ODD distribution}
\label{sec:dataset:odd}

In addition to ensuring that training, validation, and test splits are geographically disjoint, it is critical to maintain a similar distribution of selected Operational Design Domains (ODDs) across these splits. We consider a range of conditions such as road type, solar elevation, and speed, and ensure that all dataset splits closely follow a target distribution derived from real-world driving data. 
Concretely, we find that approximately 36\% of the data belongs to the road type category motorway, 52\% urban, 8\% residential, and 4\% rural; about 84\% is collected during daytime, 9\% at night, and 7\% at twilight; and 70\% of the data is recorded on dry road surfaces, 25\% under wet conditions, and 5\% on snow or ice.
While this broad coverage reflects the overall ODD distribution, we do not analyze which subset might have the highest information density, since our primary objective is to understand how much real-world driving data is needed to achieve certain performance improvements rather than identify specific, high-value segments. We verified that this ODD distribution remains valid across all dataset sizes.

\subsection{Action Labeling}
\label{sec:dataset:action}

Since our model outputs only a single trajectory, we need navigational input commands to disambiguate driving actions such as turning and lane changes.
To retrieve these action inputs, we leverage \sensitive{vectorized map labels (i.e., lane polylines and polygons) at each point in time}, referred to as a \textit{map snapshot}.
Since this data can change across timesteps, we parse each map snapshot into \textit{trajdata}~\cite{ivanovic2023trajdata} format and impute any missing road topology information \sensitive{such as lateral lane connectivity}.
Due to issues with the temporal consistency of the localization when using lanelet matching \cite{lanelet2} based on position and orientation,
we implemented a \textit{trajdata} extension that generates diverse map-based anchor paths (DMAPs) following \cite{Naumann_2023_CVPR}.
We traverse the graph to identify when the ego vehicle is:
(1) \textit{turning} as all cases with multiple outgoing longitudinal edges, and 
(2) \textit{changing lanes} as cases where a lateral edge is traversed.

Our data preprocessing pipeline relies on auto-labeling and thus may produce a small proportion of noisy samples. 
To validate label quality, we compare local linear approximations of lanelets (turn angles) with their corresponding global approximations, discovering that the rare discrepancies often stem from peculiar or complex road layouts. 
We manually inspected these instances and found they do not significantly degrade the action labels or affect our broader scaling analyses. 
We save each action’s distance in a global reference frame (driven distance from session start), since multiple snapshots can share the same action, we use majority voting to keep a single action input per snapshot. 
Ultimately, the selected data consists of 91.8\% lane keeping, 5.2\% turns, and 3\% lane changes.
More details on data quality analysis can be found in \cref{appendix:subsec:data_quality}.

While a more sophisticated action design (e.g., additional high-level behavior classes) might enhance final performance, we believe it would not significantly impact the scaling characteristics (i.e. the scaling coefficient), as the improvement will be observed across \textit{all} data regimes. Our primary focus remains on understanding how much real-world driving data is necessary to achieve certain performance gains, rather than exhaustively refining action labeling or design choices.

%% file: sec/4_methods.tex
\section{Model Architecture}
\label{sec:model}

Our model consists of two main components: a perception module that encodes camera images and a prediction module that generates future trajectories. While more advanced, state-of-the-art architectures could potentially improve final performance, our emphasis lies in the methodology and procedures for conducting systematic scaling law analyses on a representative data-driven model rather than in achieving the absolute highest accuracy. In particular, we purposely exclude past history as it can overwhelmingly indicate future states~\cite{li2024ego}, ensuring that our model learns to interpret the environment primarily from the current visual input. We believe this simpler design choice does not alter the fundamental scaling dynamics, makes our analysis more generalizable and allows us to run experiments more efficiently. For details beyond \cref{sec:model:perception,sec:model:prediction}, please refer to the Appendix.

\subsection{Perception Module}
\label{sec:model:perception}

The perception module takes as input rectified multi-view camera images. Let the rectified image from each camera view be represented by $\mathbf{I}_{v} \in \mathbb{R}^{H \times W \times 3}$, where $v \in \{f, l, r\}$ indicates the view (front, left, and right), and $H$ and $W$ represent the image height and width, respectively. The raw images are first passed through a ResNet-based encoder $\mathcal{E}(\cdot)$~\cite{HeZhangEtAl2016}, which includes a global average pooling layer that produces a single feature vector per view $\mathbf{F}_{v} = \mathcal{E}(\mathbf{I}_{v}) \in \mathbb{R}^{d}$
where $\mathbf{F}_{v}$ denotes the encoded feature vector for view $v$, and $d$ is the feature dimension obtained after the global pooling operation.

\textbf{Single-Camera Input.}
If only a single front-facing camera view is provided, the perception module directly forwards the extracted features, $\mathbf{F}_{f}$, to the prediction module for trajectory generation. 

\textbf{Multi-Camera Fusion.}
When multiple camera views are available, we fuse the multi-view information using cross attention, similar to \cite{popov2024mitigatingcovariateshiftimitation}. This fusion strategy ensures that lateral views (left and right) are integrated without bias or dependence on processing order.

\subsection{Prediction Module}
\label{sec:model:prediction}

The prediction module builds on the features generated from the perception module by incorporating additional action commands and kinematic information to produce a final trajectory output.
\vspace{-0.25cm}
\subsubsection{Action Encoding.}
The action inputs 
are represented as a tuple $(a, d, \theta)$, with:

\textbf{Action type $a$}: The type of driving maneuver, i.e. \textit{lane change left}, \textit{lane change right}, or \textit{turn}. This categorical variable is encoded using a one-hot vector, $\mathbf{A} \in \mathbb{R}^{N_a}$, where $N_a = 3$ represents the three maneuver types. Lane keeping is implicitly indicated when all three entries in the vector are set to 0, representing the absence of a specific maneuver.

\textbf{Action distance $d$}: The distance in meters from the ego's current position to where the action will occur.
It is discretized and encoded as a one-hot vector, $\mathbf{D} \in \mathbb{R}^{N_d}$, where $N_d$ represents the number of distance intervals. Each entry in $\mathbf{D}$ indicates a specific distance range.

\textbf{Action angle $\theta$}: Is the angle associated with the turn action, in degrees, and $0$ for other actions.
We represent it using both its sine and cosine values, to ensure invariance to periodicity. The encoded angle is computed as $\mathbf{\Theta} =  [\sin(\theta); \cos(\theta)]\in \mathbb{R}^2$.

The final action feature vector, $\mathbf{F}_{\text{action}}$, is obtained by concatenating the distance encoding $\mathbf{D}$, action type encoding $\mathbf{A}$, and angle encoding $\mathbf{\Theta}$, followed by passing them through a multi-layer perceptron, $\text{MLP}_{\text{action}}$, to obtain a latent feature representation: $\mathbf{F}_{\text{action}} = \text{MLP}_{\text{action}} \left( \left[ \mathbf{D}; \mathbf{A}; \mathbf{\Theta} \right] \right)$.

\subsubsection{Kinematic Encoding.}
The kinematic inputs provide the vehicle’s current state, including speed $v$, acceleration $a$, and jerk $j$. These values are represented as a feature vector $\mathbf{K} = [v, a, j] \in \mathbb{R}^3$. The kinematic feature vector is passed through $\text{MLP}_{\text{kin}}$ to encode it into a latent representation: $\mathbf{F}_{\text{kin}} = \text{MLP}_{\text{kin}}(\mathbf{K})$.

\subsubsection{Fusion and Trajectory Prediction.}
After encoding the perception, action, and kinematic features, we fuse them using cross-attention to capture dependencies between these modalities. Let $\mathbf{F}_{\text{img}}$ represent the fused image features from the perception module. The cross-attention mechanism updates $\mathbf{F}_{\text{img}}$ by attending to $\mathbf{F}_{\text{action}}$ and $\mathbf{F}_{\text{kin}}$, producing an integrated feature representation $\mathbf{F}_{\text{fused}} = \text{CrossAttn}(\mathbf{F}_{\text{img}}, [\mathbf{F}_{\text{action}}, \mathbf{F}_{\text{kin}}])$.
The final fused representation $\mathbf{F}_{\text{fused}}$ is then passed through an additional $\text{MLP}_{\text{traj}}$ to regress the predicted trajectory waypoints $\mathbf{T} = \{(x_t, y_t)\}_{t=1}^{T}$, where $T$ is the prediction horizon. In our work, $T = 15$, representing a 3-second future horizon at a frequency of 5 Hz, $\mathbf{T} = \text{MLP}_{\text{traj}}(\mathbf{F}_{\text{fused}})$.
Each trajectory point $(x_t, y_t)$ represents the vehicle’s predicted 2D position at time $t$.

\section{Training Details and Methodology}
\label{sec:training}

\subsection{Learning Rate Scheduling}
\label{subsec:challeng_training}

We employ a cosine annealing schedule to promote smooth and efficient learning \cite{kaplan2020scaling}. 
However, this requires knowing the total training duration in advance, as the cycle length must be predefined. 
In contrast to prior work that either fixes the total compute or the number of epochs per dataset size~\cite{hagele2024scaling}, we propose a linearly decreasing compute schedule. Specifically, a dataset with \(2^k\) hours of driving data is trained for \(m \times (1+\ell-k)\) epochs, where \(m=2\) represents the base number of epochs and \(\ell=13\) is the exponent \(k\) of the largest dataset split. This approach balances limited compute resources with effective training.

\begin{figure}[htbp]
    \centering
    \includegraphics[width=\linewidth]{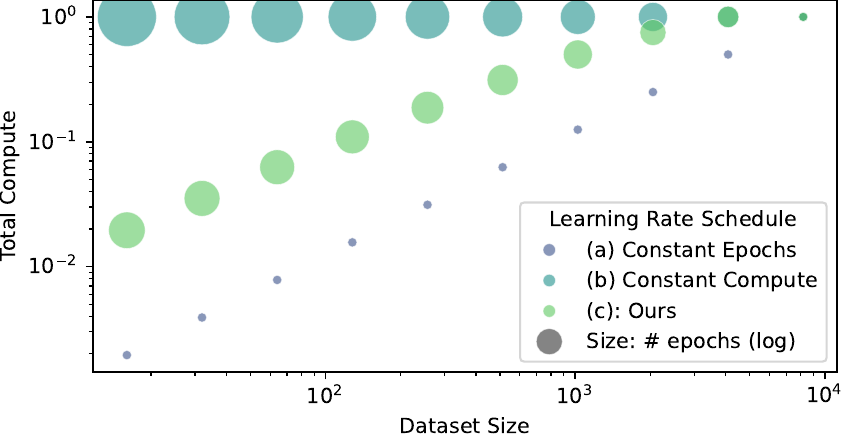}
    \caption{
    Comparison of learning rate schedules: Relationship between the total compute resources required and the dataset size. 
    }
    \label{fig:lr_schedules}
\end{figure}

As illustrated in \cref{fig:lr_schedules}, the constant epochs approach (a) applies the same number of epochs regardless of dataset size, resulting in substantial increases in compute requirements for larger datasets. In contrast, the constant compute approach (b) maintains fixed compute across varying dataset sizes, leading to a sufficient number of iterations for larger datasets but excessive compute on smaller splits. The additional requirement of a cool-down phase also makes it more complicated than cosine annealing~\cite{hagele2024scaling}. 
Our adaptive approach (c) scales the number of epochs with dataset size, providing sufficient training without over-allocating resources for smaller datasets, while still ensuring adequate resources for larger datasets. This gradual scaling achieves smooth convergence across varying dataset sizes, providing a way to calculate the cycle length for the use of cosine annealing~\cite{kaplan2020scaling}.

\subsection{Selecting Scaling Law Estimators}
\label{subsec:selecting_estimators}

\begin{figure}[htbp]
    \centering
    \begin{subfigure}{0.22\textwidth}
        \centering
        \includegraphics[width=\linewidth]{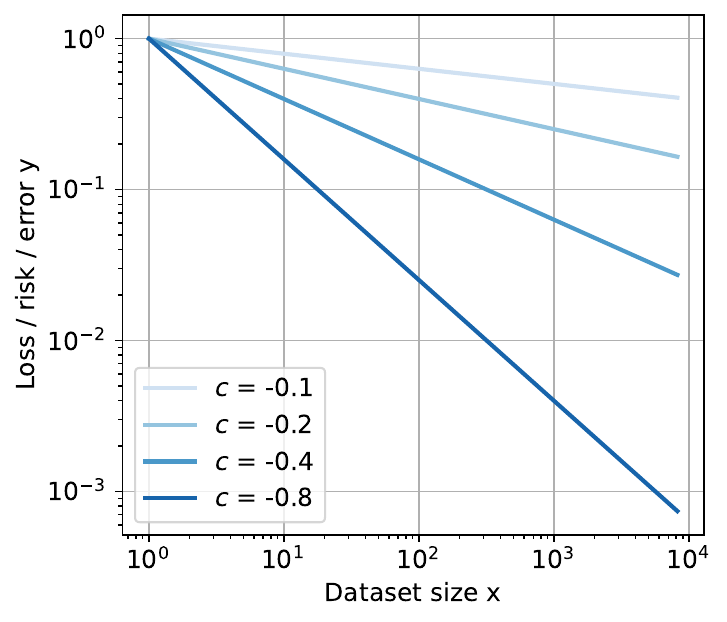}
        \caption{\textbf{M1}: \( y = \beta x^c \)}
    \label{fig:estimators:M1}
    \end{subfigure}
    \begin{subfigure}{0.22\textwidth}
        \centering
        \includegraphics[width=\linewidth]{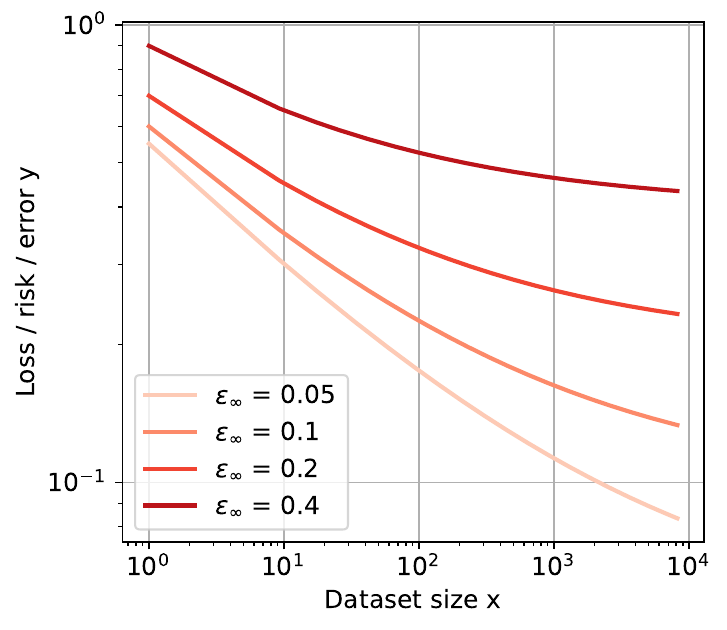}
        \caption{\textbf{M2}: \( y - \epsilon_\infty = \beta x^c \)}
    \label{fig:estimators:M2}
    \end{subfigure}
    \vskip\baselineskip  %
    \begin{subfigure}{0.22\textwidth}
        \centering
        \includegraphics[width=\linewidth]{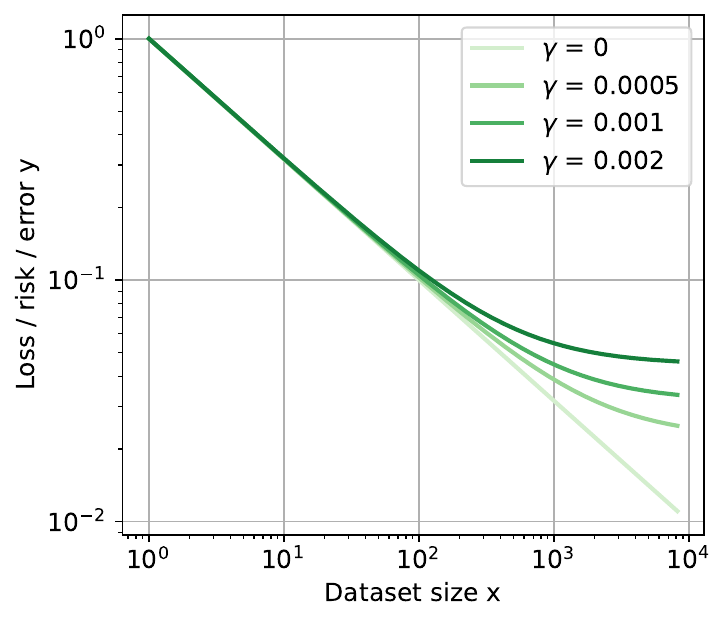}
        \caption{\textbf{M3}: \( y = \beta (x^{-1} + \gamma)^c \)}
    \label{fig:estimators:M3}
    \end{subfigure}
    \begin{subfigure}{0.232\textwidth}
        \centering
        \includegraphics[width=\linewidth]{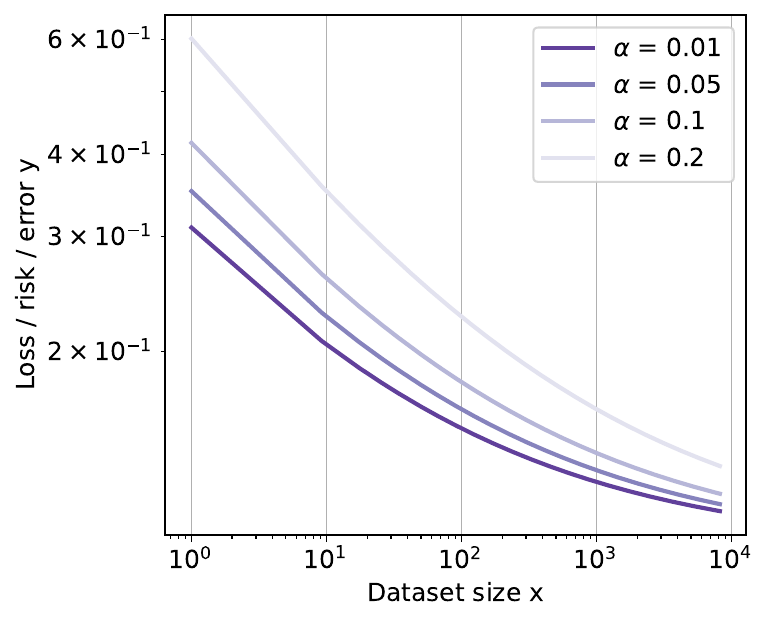}
        \caption{\textbf{M4}: \( y - \epsilon_\infty = (\epsilon_0 - y)^\alpha \beta x^c \)}
    \label{fig:estimators:M4}
    \end{subfigure}
    \vspace{-0.1cm}
    \caption{
    The scaling law estimators considered in this work, showing how specific parameters influence the resulting trend. 
    }
    \vspace{-0.4cm}
    \label{fig:estimators}
\end{figure}

Scaling law estimators generally take the form $y = f(x; \theta)$ where $y$ represents the loss, error, or certain metrics on the test set, \( x \) denotes the training dataset size, and \( \theta \) refers to the estimator parameters. In this work, we evaluate four candidate scaling law estimators—denoted as \textbf{M1}, \textbf{M2}, \textbf{M3}, and \textbf{M4}—each with increasing levels of complexity that are visualized in \cref{fig:estimators}.

\textbf{M1} is the simplest power law estimator: \( y = \beta x^c \), where \( \beta > 0 \) acts as a scaling factor and \( c < 0 \) is the exponent, governing the rate at which the loss decreases as dataset size increases \cite{kaplan2020scaling, bahri2024explaining}. 

\textbf{M2} extends M1 by introducing an offset, expressed as \( y - \epsilon_\infty = \beta x^c \), where \( \epsilon_\infty \geq 0 \) represents the asymptotic minimum loss achievable with infinite data \cite{rosenfeld2019constructive, hestness2017deep,gordon-etal-2021-data}. 

\textbf{M3} includes a shift term, resulting in the form \( y = \beta (x^{-1} + \gamma)^c \). The inclusion of \( \gamma > 0 \) allows M3 to model different scaling behaviors across small and large datasets~\cite{bansal2022data, zhai2022scaling}.

\textbf{M4} builds upon M2 by incorporating an additional lower bound and adaptability. It is defined as \( y - \epsilon_\infty = (\epsilon_0 - y)^\alpha \beta x^c \), where \( \epsilon_0 > \epsilon_\infty > 0 \) denotes the initial untrained model loss, and \( \alpha > 0 \) is an exponent that enables the function to adapt to various deviations from strict power law behavior \cite{alabdulmohsin2022revisiting}. 

Note that our primary goal is not only to find an estimator that fits the observed data points well, but also one that extrapolates beyond the given training dataset sizes. 
To this end, following \cite{alabdulmohsin2022revisiting}, we fit each estimator to the first 6 training dataset sizes (16h to 512h) and evaluate them on the next 2 training dataset sizes (1024h and 2048h)
measuring accuracy with the mean squared error (MSE) of the model's final displacement error.
For all experiments, we fit all estimators to each scenario (i.e., overall, lane changing, turning, etc.) following the above procedure to derive the most accurate scaling relationship for each action type.

%% file: sec/5_experiments.tex
\section{Experiments}
\label{sec:experiments}

Our experiment setup is detailed in \cref{sec:experiment_setup}, with open-loop and closed-loop results in \cref{sec:openloop,sec:closedloop}, respectively.

\subsection{Experiment Setup}
\label{sec:experiment_setup}

\textbf{Metrics.}
For trajectory prediction evaluation, we use standard metrics widely adopted in recent prediction challenges~\cite{ChangLambertEtAl2019,wilson2021argoverse2,waymo_open_motion_dataset}: Average Displacement Error (ADE), Final Displacement Error (FDE), and Miss Rate (MR)~\cite{ChangLambertEtAl2019}. Unlike previous works that output multiple potential trajectories, our model produces one single output trajectory. 
The ADE metric computes the average Euclidean (\(\ell_2\)) distance in meters over all future time steps between the predicted trajectory and the ground truth. FDE calculates the Euclidean distance at the final predicted time step, providing an assessment of endpoint accuracy. MR measures the proportion of scenarios where the endpoint of the predicted trajectory deviates from the ground truth by more than $2.0$ meters, indicating instances where the prediction misses the target significantly.

\textbf{Implementation Details.}
 We transform each scene into an ego-centric coordinate frame, focusing solely on predicting the trajectory of the ego vehicle. The learning rate follows a cosine annealing schedule as outlined in \cref{subsec:challeng_training}. Further details on the training procedure and runtime can be found in \cref{appendix:section:experiments}.

\textbf{Model Size.}
To investigate the effect of scaling model sizes, we varied the backbone of the perception module by using two different image encoders: ResNet-18 (RN18) with 11.2 million parameters and ResNet-50 (RN50) with 24.3 million parameters. The prediction module, containing 7.2 million parameters, was kept constant across all setups to ensure consistency. %

\subsection{Open-Loop Evaluation}
\label{sec:openloop}

\subsubsection{ResNet-18 with 1 Camera}
\label{sec:res:rn18_1cam}

\textbf{Extrapolation performance.}
We first examine the prediction performance across all dataset splits, using FDE as the primary indicator. To determine the best scaling estimator, we follow the selection process outlined in \cref{subsec:selecting_estimators}, and identify that \textbf{M2} provides the best fit for overall performance, lane keeping, and lane changing, achieving the lowest MSE on the 1024-hour and 2048-hour validation splits. For turning scenarios, however, \textbf{M3} offers the most accurate fit. Each estimator is then retrained, incorporating all points up to 2048 hours, and evaluated on the 4096- and 8192-hour splits. To evaluate estimator performance, we compare the predicted dataset sizes required to reach FDE levels observed with the 4096-hour and 8192-hour training splits to the actual dataset sizes, as shown in \cref{table:m2_examples}. Overall, we observe that both estimators tend to be optimistic in their required dataset size predictions, particularly for all actions combined, lane keeping, and turning scenarios. Lane changing, however, shows closer alignment between predicted and actual dataset sizes, indicating more accurate estimates for this action type. We acknowledge this optimistic tendency in predictions and will discuss it further as a limitation in the conclusion. 

\begin{table}[h!]
\centering
\resizebox{0.9\columnwidth}{!}{
\begin{tabular}{cccr}
       \textbf{RN18 - 1 Cam}               &  &\multicolumn{2}{c}{\textbf{Dataset Size (h)}} \\ 
\textbf{Action Type} & \textbf{Target FDE}     & \textbf{Actual }         & \textbf{Pred. }    \\ \hline
All [M2]              & 0.936     & 4,096   & 2,593~$\downarrow$              \\
All [M2]              & 0.912     & 8,192   & 4,793~$\downarrow$             \\
None (Lane keeping) [M2]              & 0.903     & 4,096   & 2,609~$\downarrow$             \\
None (Lane keeping) [M2]              & 0.880     & 8,192   & 4,931~$\downarrow$             \\
Lane change [M2]& 1.583 & 4,096 & 2,879~$\downarrow$ \\
Lane change [M2]& 1.543 & 8,192 & 10,000~$\uparrow$ \\
Turning [M3]& 1.793 & 4,096 & 1,961~$\downarrow$\\
Turning [M3]& 1.731 & 8,192 & 2,789~$\downarrow$ \\
\end{tabular}}
\caption{
    Comparison of the estimator's dataset size predictions with the actual dataset sizes for a given target performance.
}
\vspace{-0.2cm}
\label{table:m2_examples}
\end{table}

In the rest of this subsection, we include all the observation points (from 16h to 8192h) for training the estimator in our analyses.

\begin{figure}[t]
    \centering
    \includegraphics[width=0.8\linewidth]{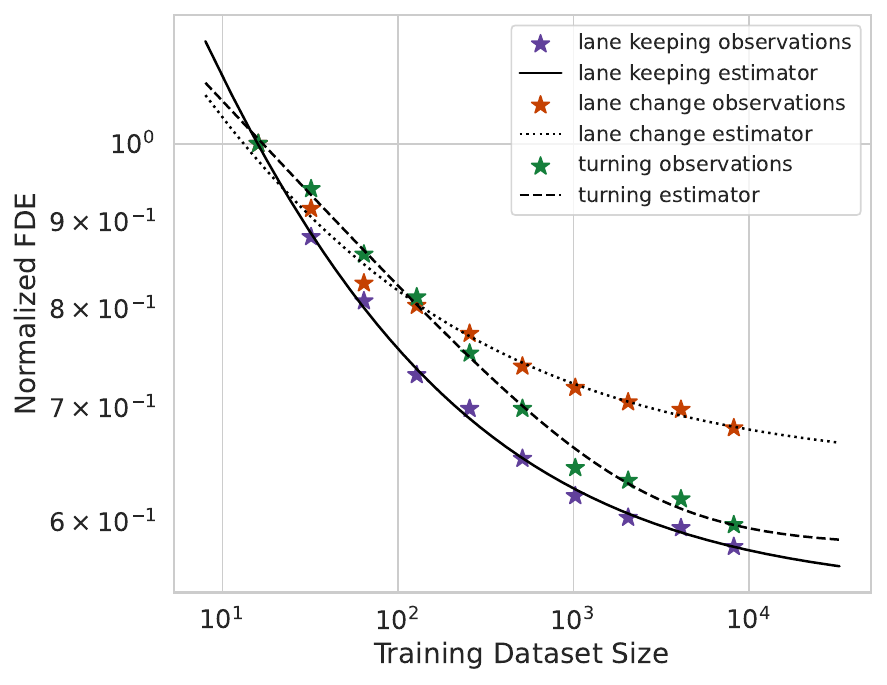}
    \caption{Relative comparison of improvements in lane keeping, lane changes and taking turns with increasing training set size.
    }
    \label{fig:compare_driving_scenarios}
    \vspace{-0.3cm}
\end{figure}

\textbf{Do all action types scale at the same rate?}
To directly compare the scaling behaviors of lane keeping, lane changing, and turning, we normalize their FDE values such that the highest value for each action type is set to 1.0, as shown in \cref{fig:compare_driving_scenarios}. 
This normalization enables a unified view of how each scenario benefits from increasing dataset size. 
We fit \textbf{M2} to lane keeping and lane change, \textbf{M3} to turning, with the resulting parameter values presented in \cref{table:compare_driving_scenarios_m2_params}.
Lane keeping shows the most substantial improvement with dataset scaling, reflected by its high slope coefficient $c$. In contrast, lane changing demonstrates a slower improvement, indicating diminishing returns from additional data. 
For turning scenarios, we observe a consistent scaling relationship up to 8192 hours, after which a plateau begins to emerge.

\begin{table}[h!]
\centering
\resizebox{0.8\columnwidth}{!}{
\begin{tabular}{l|ccc}
 \textbf{RN18 - 1 Cam}                     & \multicolumn{3}{c}{\textbf{Estimator Parameters}} \\ \cline{2-4} 
\textbf{Action Type} & \bm{\beta}     & \bm{c}         & \bm{\epsilon_{\infty}} / $\gamma$    \\ \hline
None (Lane keeping) [M2]   & 1.422     & -0.413    & 0.5457              \\
Lane change [M2]         & 0.873     & -0.348    & 0.6444              \\
Turning [M3]              & 1.365     & 0.110    & 0.0004                 
\end{tabular}}
\caption{
Parameters of the estimators for different actions fitted on all ten observations (from 16h to 8192h).
}
\label{table:compare_driving_scenarios_m2_params}
\end{table}

\begin{figure}[t]
    \centering
    \includegraphics[width=0.8\linewidth]{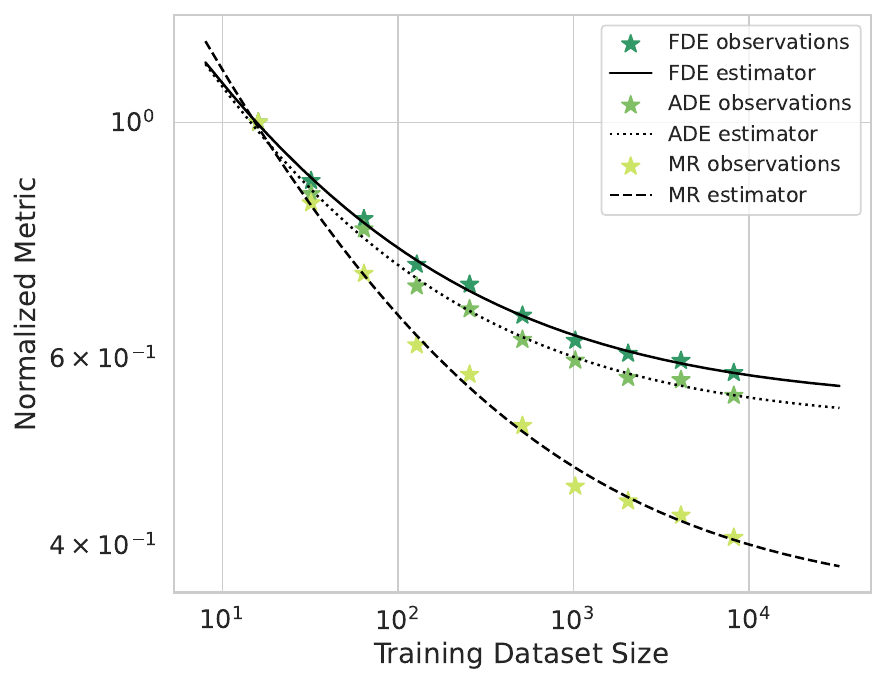}
    \vspace{-0.1cm}
    \caption{
    Scaling law analysis comparing FDE, ADE and MR$_{2m}$.
    }
    \label{fig:compare_metrics}
    \vspace{-0.3cm}
\end{figure}

\vspace{-0.1cm}
\textbf{Do all metrics scale at the same rate?}
In \cref{fig:compare_metrics}, we examine additional open-loop metrics, ADE and MR, alongside FDE to compare their scaling behaviors. After a similar estimator selection process as in \cref{subsec:selecting_estimators}, we find that \textbf{M2} provides the best fit across all three metrics. For consistency in the comparison, each metric is normalized such that its highest value is mapped to $1.0$. As can be seen in \cref{table:different_metrics}, all metrics follow similar scaling trends with scaling coefficients $c$ around $-0.4$ across metrics. However, despite the close values of $c$, MR$_{2m}$ displays a more pronounced decline compared to FDE and ADE. This is partly due to its smaller offset parameter, \( \epsilon_{\infty} = 0.352 \), which allows it to more quickly achieve lower values. This behavior is consistent with intuition, as MR$_{2m}$ is a discrete threshold-based metric and tends to register improvement more abruptly once prediction endpoints fall within the 2-meter threshold.

\begin{table}[t]
\centering
\resizebox{0.5\columnwidth}{!}{
\begin{tabular}{c|ccc}
                 & \multicolumn{3}{c}{\textbf{M2 Parameters}} \\ \cline{2-4} 
\textbf{Metrics} & $\beta$     & $c$          & $\epsilon_{\infty}$   \\ \hline
FDE              & 1.358    & -0.396    & 0.543            \\
ADE              & 1.464    & -0.417    & 0.520            \\
MR$_{2m}$         & 1.925    & -0.399    & 0.352           
\end{tabular}}
\caption{M2 parameters trained on normalized metrics.}
\label{table:different_metrics}
\end{table}

\textbf{Extrapolation Beyond Our Data.}
Using all data up to the 8192-hour split as training set, we extend our scaling law analysis to predict the additional dataset size required to improve upon the final FDE achieved at this level. Results show that a 1\% improvement in FDE would require approximately 4,000 hours of additional driving data, while a 3\% improvement demands around 29,000 hours. 
A 5\% improvement, furthermore, would necessitate an extensive 273,000 hours. This exponential growth in data requirements emphasizes diminishing returns even if we scale up training data, especially considering that both the \textbf{M2} and \textbf{M3} estimators have already exhibited slightly optimistic predictions.

\vspace{-0.3cm}
\subsubsection{ResNet-50 with 1 Camera}
After observing diminishing returns in performance with increasing data in prior experiments, we hypothesized that our model’s limited capacity might be a contributing factor. To investigate this, we analyzed the scaling law using a larger ResNet-50 backbone, increasing the backbone size from 11.2M to 24.3M parameters and raising the total model size from 18.4M to 31.5M parameters. This aligns with trends in NLP, where model size, compute, and dataset size are scaled in tandem to achieve improvements.

With the ResNet-50 backbone, we observe significant improvements in scaling across all scenarios (cf. \cref{fig:intro}). For overall performance, the ResNet-50 model achieves the same FDE with only around 3000 hours of data—a 63\% reduction compared to the original 8192 hours required by ResNet-18. Similarly, lane keeping requires 2970 hours (64\% reduction), lane changing 1815 hours (78\% reduction), and turning 2597 hours (68\% reduction) to reach comparable FDE values.

These findings highlight the critical role of model capacity in scaling efficiency. By increasing the model size, we achieve faster performance gains across all scenarios, effectively reducing the data requirements for comparable levels of accuracy. This highlights the importance of scaling model capacity alongside dataset size for end-to-end autonomous driving systems.

\vspace{-0.3cm}
\subsubsection{ResNet-18 with 3 Cameras}

To further explore the effects of input diversity, we extended our scaling law experiments by adding three camera inputs (front, left, and right) to the model, while keeping the ResNet-18 backbone and prediction module configuration. Tested on datasets from 16 to 1024 hours, we found that the overall FDE with three cameras remained comparable to that of a single-camera setup. We attribute this similarity to the dataset composition, which predominantly features lane-keeping scenarios.

A detailed analysis on the 512-hour split revealed that while FDE for lane keeping remained virtually unchanged between one and three cameras, the multi-camera setup led to a modest 4-5\% improvement in more complex actions such as lane changing and turning. This suggests that additional camera inputs offer greater value in complex maneuvers but have a limited impact on overall performance due to the dominance of lane-keeping in the dataset.

\subsection{Closed-Loop Evaluation}
\label{sec:closedloop}

Finally, we evaluate the closed-loop performance of our models in \drivesim{}, a high-fidelity simulator with enhanced photorealism and realistic physics models. While real-vehicle testing is ideal, it requires extensive engineering effort and resources, placing it outside our present scope. Instead, we leverage \drivesim{} as it has been validated in prior works \cite{popov2024mitigatingcovariateshiftimitation}, providing a more challenging environment than other popular simulators, e.g., CARLA. 

In our experiments, we focus on two highway scenarios with minimal traffic interactions, where the primary challenge is lane-keeping. We measure the MDBF, defined as the distance driven before the vehicle departs the drivable area. We transfer two models---trained with ResNet-18 and ResNet-50 backbones on different data subsets---into \drivesim{}. Despite observing consistent open-loop performance gains at larger data scales, the MDBF results improved only up to the 256-hour mark, beyond which they plateaued around 1\,km. This early plateauing phenomenon, in contrast to open-loop scaling, has also been observed in the closed-loop experiments of \cite{zheng2024preliminarydatascaling}.

To address potential real-to-sim discrepancies, we applied image augmentations during training to better align real-world visuals with simulator conditions. We also explored increasing model capacity (planner head dimensions of 256, 512, and 1024) with a ResNet-18 backbone. However, neither data augmentation nor capacity scaling alleviated the plateau in closed-loop performance.

This observation reinforces findings from prior research: open-loop improvements do not directly translate to closed-loop performance~\cite{dauner2023parting}. Although open-loop metrics continue to improve with larger data, these gains fail to materialize proportional closed-loop benefits. This disconnect is partly due to “covariate shift” in imitation learning, where models encounter untrained states during deployment. Techniques like DAgger (Dataset Aggregation)~\cite{ross2011reduction} can help address this by incorporating corrective data into training. This includes generating recovery trajectories to guide the vehicle back to the lane center during near-failure states and retraining the model with this augmented data. Alternatively, integrating auxiliary tasks to enhance situational reasoning could further bridge this gap. Finally, moving beyond open-loop behavior cloning towards closed-loop training~\cite{popov2024mitigatingcovariateshiftimitation} is a promising strategy as it much more closely resembles real-world driving.

Although we could not conduct large-scale real-world tests, our limited closed-loop results do suggest a transferable trend up to 256\,hours, beyond which performance saturates in the chosen scenario. Future work may systematically explore longer time scales, more diverse scenarios, and advanced data-collection protocols to further bridge the sim-to-real gap.

%% file: sec/6_conclusion.tex
\section{Conclusions}
\label{sec:conclusion}

In this work, we investigate the scaling performance of a simple end-to-end autonomous driving system in both open- and closed-loop settings across various dataset sizes. We identify several key findings: (1) Effective scaling in end-to-end driving requires concurrent increases in both dataset size and model capacity; (2) Scaling dynamics vary significantly across tasks (e.g., lane-keeping, lane-changing, turning), emphasizing the need for targeted data collection and curation; (3) While single-camera input performs adequately for lane-keeping tasks, surround-view inputs are essential for more complex scenarios such as lane-changing and turning; (4) Open-loop performance does not directly translate to closed-loop driving capabilities. 

Several areas for further exploration remain. Integrating additional sensor modalities, such as LiDAR, radar, or additional (e.g., rear-facing) cameras could improve the handling of complex scenarios. Incorporating temporal information and multi-agent dynamics would align the analysis more closely with state-of-the-art AV stacks. Using more accurate action labeling methods, such as human annotations, could reduce dataset noise, resulting in more reliable scaling laws. Extending data scaling analysis to advanced end-to-end models, such as UniAD~\cite{jiang2023vad} or PARA-Drive~\cite{weng2024drive}, may reveal which modules enhance prediction accuracy the most, especially in complex scenarios such as unprotected turns through an intersection. Finally, while our estimator demonstrates strong fitting capabilities, we acknowledge its optimistic tendencies in out-of-sample predictions and consider enhancing its predictive accuracy a key direction for future work.

%% file: sec/X_suppl.tex
\clearpage
\setcounter{page}{1}
\maketitlesupplementary
\appendix

\section{More Details on the Dataset}

\begin{figure}[h]
   \centering
   \includegraphics[width=0.95\linewidth]{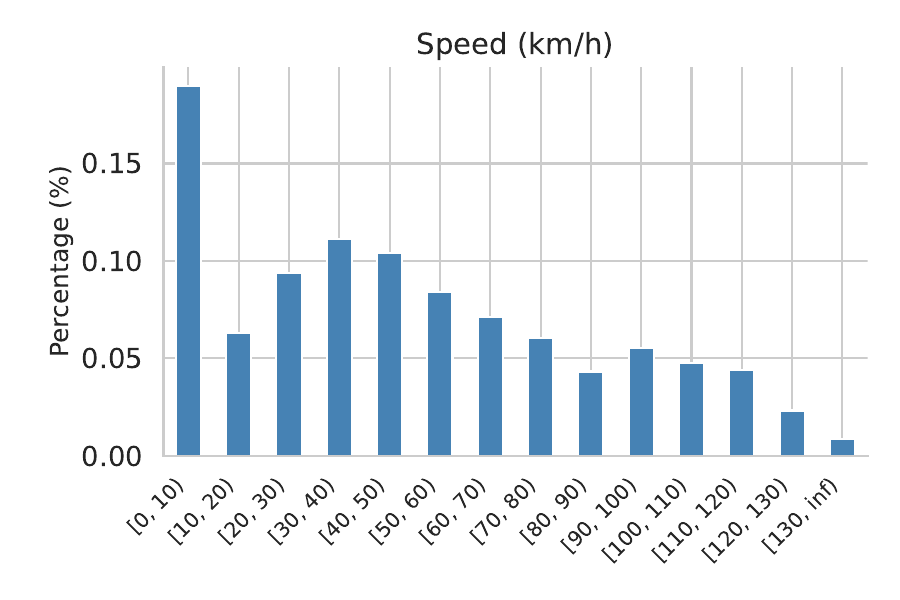}
   \caption{Speed distribution of the ego vehicle.}
   \label{fig:speed}
\end{figure}

\begin{figure}[h]
   \centering
   \includegraphics[width=0.95\linewidth]{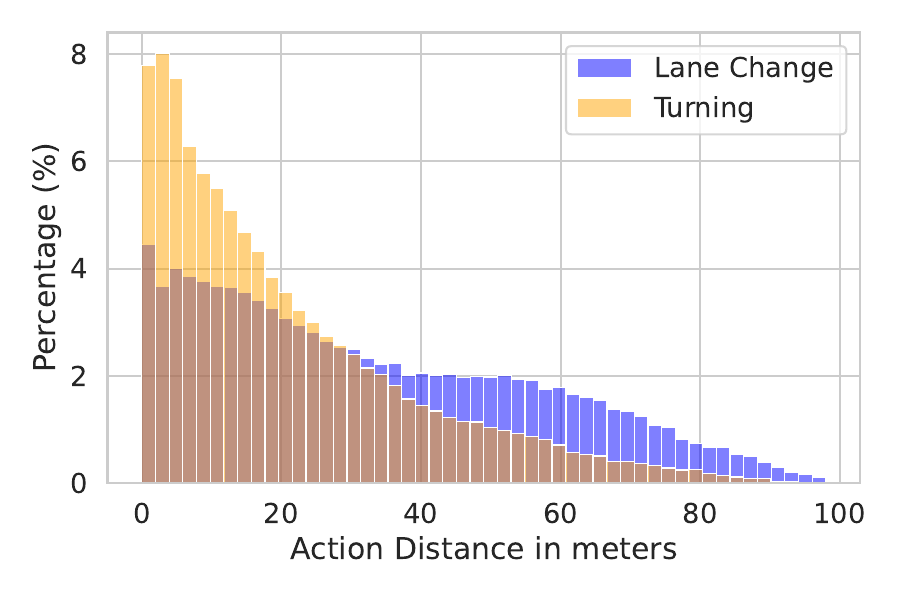}
   \caption{Distribution of the action distance.}
   \label{fig:action_distance}
\end{figure}

We analyze the distribution of the vehicle's speed and the distance to take actions (i.e. lane changing and taking a turn). We group these statistics into bins with an interval of 10 km/h for the speed and 2 m for the action distance. The distributions are provided in \cref{fig:speed} and \cref{fig:action_distance} respectively.

\subsection{Data Quality Analysis}
\label{appendix:subsec:data_quality}
\begin{figure}[h]
   \centering
   \includegraphics[width=0.95\linewidth]{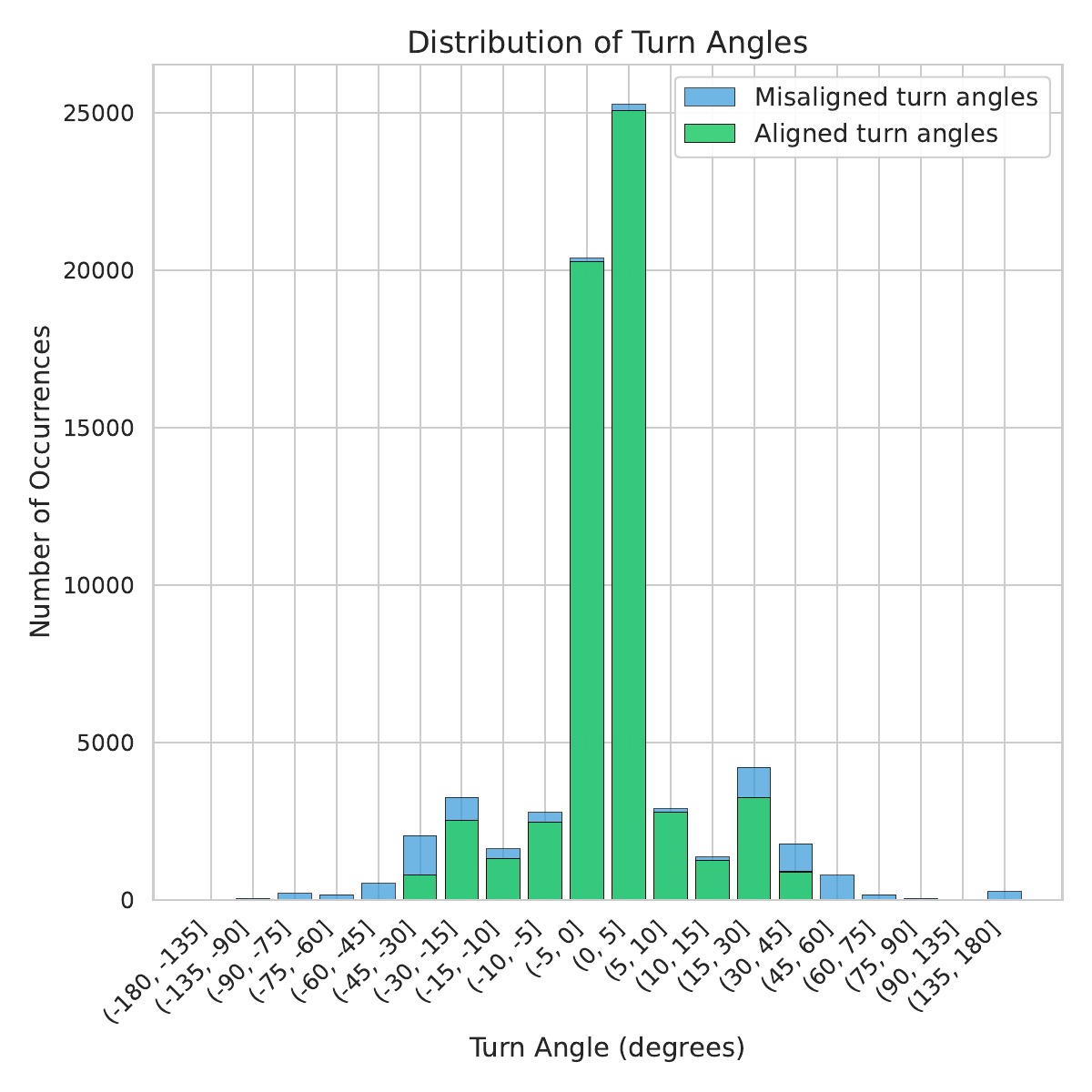}
   \caption{Turn angle alignment.}
   \label{fig:action}
\end{figure}

Our data pre-processing pipeline relies on auto-labeling, which may result in some portion of noisy samples i.e. with inaccurate turn angles. We validated label quality by comparing turn angles (local linear approximations of lanelets) with their global counterpart (linear approximations of full lanelets).
\cref{fig:action} (left) shows where they match (green) and where they don't (red) and \cref{fig:action} (right) shows the ratio of accurately labeled turn angles for different angle bins and we show that the accuracy of our labeling drops in sharper turns.

\subsection{Data Curation Procedure}

We provide some additional information for the dataset curation pipeline in the following.

\textbf{Geofencing. }
We utilize an \sensitive{H3 cell resolution of 11, corresponding to a cell edge length of 24.91m and a cell diameter of 49.82m}.

\textbf{ODD Distribution. }
Maintaining near-identical ODD proportions in splits smaller than 32 hours remains challenging, especially when factoring in large session clusters. We investigated breaking up these clusters by analyzing which H3 cells are responsible for large connected regions and assessing what would happen if they were removed, but ultimately opted against this strategy because the cluster connections were too complex and dense in our data. Instead, to obtain the final splits, we iteratively grow each dataset portion by randomly sampling a session cluster from the smallest 50\% of the remaining ones and assigning it to the split that best preserves the desired distribution. Repeating this approach multiple times and selecting the configuration closest to the real-world distribution yielded the best results, and we verified that this ODD distribution remains valid across all dataset sizes. 

\textbf{Action Labeling. }
Due to issues with the temporal consistency of the localization when using lanelet matching \cite{lanelet2} based on position and orientation,
we implemented a \textit{trajdata} extension that generates diverse map-based anchor paths (DMAPs) following \cite{Naumann_2023_CVPR}.
This enables the computation of map-based anchor paths for the ego-vehicle from any initial position.
For each anchor path, we measure alignment with the ground truth ego-motion to determine the ego's future lane sequence.
The matching score is computed as 
\[
s = \alpha  s_{\text{IoU}} + (1 - \alpha) s_{\text{LI}}
\]
where $s_{\text{IoU}}$ is the Intersection over Union (IoU) of the ego and anchor paths (each with 1m of buffer), $s_{\text{LI}}$ is the percentage of the ego path within the lanelets given by the anchor path, and weighting factor $\alpha \in [0,1]$.
We save the action distance in a global reference frame, i.e. as driven distance from the session start.
Since multiple snapshots will contain the same action, this global reference frame enables the removal of noisy snapshot data by using majority voting to determine the final action.
To simplify the action encoding for training, we only take actions within the ground truth traveled distance into account and save only one action conditioning input per snapshot, although multiple inputs might have been generated.
We use manual visual debugging to verify the procedure qualitatively.

\subsection{Corner Case Handling} 

We focus on analyzing overall model performance improvement as dataset size increases and provide a more comprehensive understanding of AV data-scaling laws, rather than analyzing corner cases. 
Detecting corner cases would require (subjective) definitions,  targeted labeling, and additional mechanisms like novelty or out-of-distribution detection, which are out of the scope of this work.

\section{Model Diagram and Details}

We provide details on our perception module in \cref{app:model:perception}.

\subsection{Details on the Perception Module}
\label{app:model:perception}

We provide further clarification on the variables and operations introduced in the Perception Module. 
The perception module processes input images from multiple camera views and extracts feature representations for downstream tasks. 
Below is a brief explanation of the key variables used:
\begin{itemize}
    \item \textbf{$\mathbf{I}_{v} \in \mathbb{R}^{H \times W \times 3}$}: The rectified image from each camera view, where $v$ denotes the camera view (front $f$, left $l$, or right $r$).
    
    \item \textbf{$\mathcal{E}(\cdot)$}: The ResNet-based encoder used to extract features from each input image. It includes a global average pooling layer that reduces the spatial dimensions of the feature maps to a single vector.
    
    \item \textbf{$\mathbf{F}_{v} = \mathcal{E}(\mathbf{I}_{v}) \in \mathbb{R}^{d}$}: The encoded feature vector for each view $v$, where $d$ is the dimension of the feature vector after global pooling.
    
\end{itemize}
The multi-view fusion process leverages cross-attention to combine information from all available views (front, left, and right), ensuring balanced integration of lateral perspectives:

\begin{enumerate}
\item Define the front view’s feature map, $\mathbf{F}_f$, as the query ($\mathbf{Q}$), and use the left and right feature maps, $\mathbf{F}_{l}$ and $\mathbf{F}_{r}$, as keys ($\mathbf{K}$) and values ($\mathbf{V}$) in separate cross-attention layers.

\item Apply cross-attention to combine features, where each cross-attention layer updates $\mathbf{F}_f$ by attending to $\mathbf{F}_{l}$ and $\mathbf{F}_{r}$:
\[
\mathbf{A}_{f, l} = \text{CrossAttn}(\mathbf{F}_f, \mathbf{F}_{l})
\]
\[
\mathbf{A}_{f, r} = \text{CrossAttn}(\mathbf{F}_f, \mathbf{F}_{r})
\]

Here, $\mathbf{A}_{f, l}$ and $\mathbf{A}_{f, r}$ represent the attention outputs for the front-left and front-right interactions, respectively. These outputs are then aggregated to form the final fused representation.

\item \textbf{Symmetric Fusion}: The final fused feature map, $\mathbf{F}_{\text{fused}}$, is computed by aggregating the cross-attention outputs. We use a simple element-wise summation:
\[
\mathbf{F}_{\text{img}} = \mathbf{F}_f + \mathbf{A}_{f, l} + \mathbf{A}_{f, r}
\]
This symmetric fusion captures contextual information from both lateral views equally, enhancing the spatial awareness of the front view.
\end{enumerate}

\section{Impact of Varying Training Points on Scaling Law Estimators}

\begin{figure}[t]
    \centering
    \includegraphics[width=0.8\linewidth]{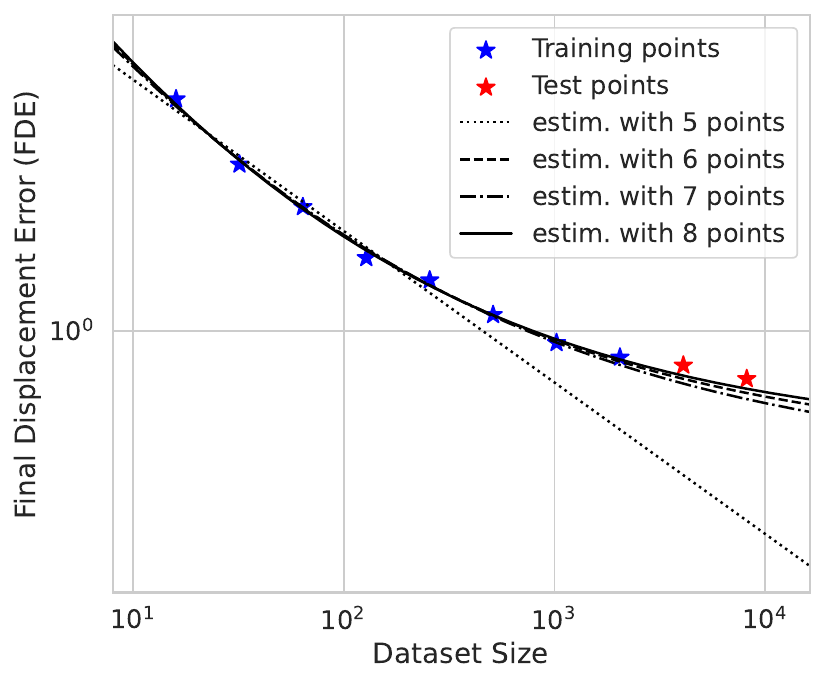}
    \caption{
    Analyzing the performance for \textbf{M2}: \( y - \epsilon_\infty = \beta x^c \) in the case of iteratively increasing the number of training points.
    }
    \label{fig:est_it}
\end{figure}

As illustrated in \cref{fig:est_it} and \cref{tab:est_it}, incorporating more data points to train the \textbf{M2} estimator on FDE values enables a closer fit to the scaling curve, improving the alignment with the data and leading to lower extrapolation loss on the test set.

\begin{figure*}[t]
    \centering
    \includegraphics[width=0.95\linewidth]{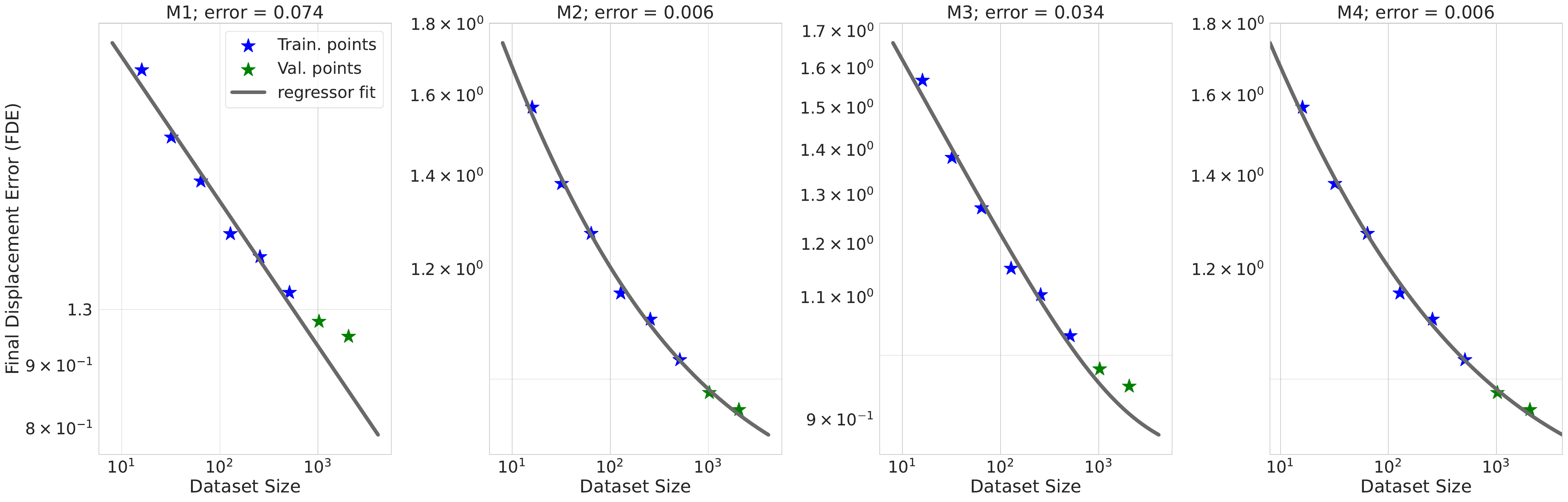}
    \caption{
    Comparison of fitting all scaling law estimators on the full dataset.
    }
    \label{fig:comp_est_overall}
    \vspace{-0.3cm}
\end{figure*}

\section{Selecting Scaling Law Estimators}

As described in the method section, the estimators are trained using the first six data points, with the 1024-hour and 2048-hour points reserved for validation. \Cref{fig:comp_est_overall} illustrates the scaling law fitting on FDE across all scenarios. Among the estimators, \textbf{M2} and \textbf{M4} demonstrate the best fit, achieving the lowest mean squared error (MSE). In contrast, \textbf{M1} fails to capture the trend and reduces to a straight line (a pure power law), while \textbf{M3} provides overly optimistic estimates, with values decreasing too quickly toward the end, resulting in a higher MSE.
Following Occam's Razor, we choose \textbf{M2} over \textbf{M4} in our analysis.

Moreover, we believe the discrepancies in scaling law curve-fitting arise more from the inherent challenges of scaling law estimation than dataset quality, as existing estimators exhibit varying levels of robustness across different data regimes. Collecting measurements averaged over multiple runs could reduce noise and improve curve-fitting, but would require significantly more compute power and time. A piecewise function could reduce error, but would introduce artificial discontinuities. Additionally, the plateau becomes apparent only for datasets exceeding $10^3$ hours, with metrics like FDE and ADE improving steadily in a near-linear trend on a log-log scale before this point. This reflects natural diminishing returns at larger scales, also seen in scaling law analyses in the computer vision and natural language domains, rather than a limitation of data quality.

\begin{table}[t]
\centering
\resizebox{0.99\columnwidth}{!}{
\begin{tabular}{|c|c|c|c|c|}
\hline
\# Points & $\beta$ & $c$ & $\epsilon_\infty$ & Extrapolation Loss \\ 
\hline
5 & 2.1837 & -0.1274 & 0.0000 & $0.2002 \pm 0.0135$ \\ 
6 & 1.9010 & -0.3319 & 0.7917 & $0.0244 \pm 0.0002$ \\ 
7 & 1.8594 & -0.3133 & 0.7663 & $0.0338 \pm 0.0009$ \\ 
8 & 1.9488 & -0.3486 & 0.8103 & $0.0179 \pm 0.0004$ \\ 
\hline
\end{tabular}}
\caption{Quantitative extrapolation results for \textbf{M2}: \( y - \epsilon_\infty = \beta x^c \) using an iteratively increasing numbers of training points, complementing the visualization in \cref{fig:est_it}.}
\label{tab:est_it}
\end{table}

\section{Experimental Setup}
\label{appendix:section:experiments}

\subsection{Hyperparameters}

We use the Adam optimizer for training our models without applying any weight decay. Training is conducted in FP32 precision, as we encountered instabilities when using FP16 or BF16 precisions. We employ a cosine annealing schedule for the learning rate, with the final learning rate set to $0$ (i.e., $\eta_{min} = 0$). The initial learning rate ($\eta_{max}$) is scaled linearly with the total effective batch size in distributed training. Specifically, we use a learning rate of $0.001$ for an effective batch size ($bs$) of $1024$, and scale the initial learning rate for other batch sizes accordingly: 

\begin{itemize}
    \item $bs=512 \rightarrow \eta_{max} = 0.0005$
    \item $bs=1024 \rightarrow \eta_{max} = 0.001$
    \item $bs=2048 \rightarrow \eta_{max} = 0.002$
    \item etc.
\end{itemize}

\subsection{Training Time and Hardware}
We use a compute cluster consisting of A-100 GPUs. Particularly, training a model with the ResNet-18 backbone on the largest data split (8192 hours) takes around 24 hours on 8$\times$A-100 GPUs.